\documentclass[runningheads]{llncs}

\usepackage[year=2024,ID=4515]{eccv}

\usepackage{eccvabbrv}

\usepackage{graphicx}
\usepackage{booktabs}

\usepackage[accsupp]{axessibility}  
\usepackage{graphicx}
\usepackage{amsmath}
\usepackage{amssymb}
\usepackage{booktabs}

\usepackage{epsfig}
\usepackage{graphicx}
\usepackage{amsmath}
\usepackage{amssymb}

\usepackage{listings}
\usepackage{booktabs}
\usepackage{color}
\usepackage{graphicx}
\usepackage{multirow}
\usepackage{colortbl}
\newcommand\blfootnote[1]{%
  \begingroup
  \renewcommand\thefootnote{}\footnote{#1}%
  \addtocounter{footnote}{-1}%
  \endgroup
}

\usepackage[pagebackref,breaklinks,colorlinks,citecolor=eccvblue]{hyperref}

\usepackage{orcidlink}

\usepackage{amsmath}
\usepackage{amssymb}

\usepackage{xcolor}
\definecolor{citecolor}{HTML}{014A98}
\hypersetup{
    colorlinks=true,
    linkcolor=red,
    filecolor=magenta,      
    urlcolor=magenta,
    citecolor=citecolor,
}

\begin{document}

\title{	
High-Precision Self-Supervised Monocular Depth Estimation with Rich-Resource Prior
}
\titlerunning{RprDepth}

\author{
Wencheng Han\orcidlink{0009-0005-2358-6969} \and
Jianbing Shen $^\dagger$\orcidlink{0000-0003-1883-2086}, 
}
\authorrunning{W. Han et al.} 
\institute{SKL-IOTSC, Computer and Information Science, University of Macau, China
}

\maketitle

\begin{abstract}
In the area of self-supervised monocular depth estimation, models that utilize rich-resource inputs, such as high-resolution and multi-frame inputs, typically achieve better performance than models that use ordinary single image input. 
However, these rich-resource inputs may not always be available, limiting the applicability of these methods in general scenarios. 
In this paper, we propose Rich-resource Prior Depth estimator (RPrDepth), which only requires single input image during the inference phase but can still produce highly accurate depth estimations comparable to rich-resource based methods.
Specifically, we treat rich-resource data as prior information and extract features from it as reference features in an offline manner.
When estimating the depth for a single-image image, we search for similar pixels from the rich-resource features and use them as prior information to estimate the depth. Experimental results demonstrate that our model outperform other single-image model and can achieve comparable or even better performance than models with rich-resource inputs, only using low-resolution single-image input. 

\textbf{Code:} \hyperlink{https://github.com/wencheng256/RPrDepth}{https://github.com/wencheng256/RPrDepth}
\end{abstract}

\blfootnote{$^{\dagger}$Corresponding author: \textit{Jianbing Shen}. This work was supported in part by the FDCT grants
0102/2023/RIA2, 0154/2022/A3, and 001/2024/SKL, the MYRG-CRG2022-00013-IOTSC-ICI grant and the SRG2022-00023-IOTSC grant.
           }
\section{Introduction}
\label{sec:intro}

Depth estimation is a crucial component in computer vision, particularly for applications like autonomous driving, where understanding the 3D structure of the environment is essential for navigation and decision-making. Traditionally, depth information has been obtained using stereo vision~\cite{liu2014optimized,laga2020survey} or LiDAR systems. However, these methods can be costly and complex, motivating the exploration of monocular depth estimation. Monocular depth estimation involves deducing the depth information of a scene from a single camera. This is inherently challenging as it requires the model to infer 3D information from 2D data, a task that humans do effortlessly but is complex for machines due to the loss of spatial information in a single image.

Recent advancements in monocular depth estimation have opened avenues for simpler, more cost-effective solutions. 
Godard \textit{et al.}~\cite{monodepth2} introduced a simplified self-supervised model for monocular depth estimation. They employ innovative loss functions and sampling methods to achieve promising depth accuracy. Subsequently, many other methods improve the performance further by designing better network architectures~\cite{hrdepth,han2023self}, using more suitable loss functions~\cite{poggi2020uncertainty,watson2019self,kumar2020unrectdepthnet,poggi2020uncertainty}.
Watson \textit{et al.}~\cite{manydepth} proposed an adaptive deep end-to-end cost volume-based method for dense depth estimation. 
Their method utilizes sequence information at test time and introduces a novel consistency loss to enhance the performance of self-supervised monocular depth estimation networks. Although this method achieves a significant improvement compared to previous works, it requires richer-resource inputs, specifically multi-frame data, during inference. Many methods follow this approach and propose highly effective depth estimators using multi-frame data inputs~\cite{guizilini2022multi,feng2022disentangling}.

In this paper, we refer to high-resolution, multi-frame data as ``rich-resource data''. We have noticed that many of the best-performing methods depend on rich-resource data. This poses significant challenges in real-world scenarios. In some situations, acquiring rich-resource inputs is impractical. For instance, multi-frame based models necessitate the capture of multi-frame data from varied positions. However, when cars are stationary, obtaining images from different positions is not possible. Moreover, many multi-frame based models demonstrate improved performance when future frames are available, but these cannot be obtained in real-world applications. Hence, there is a need for a method that can generate a comparable depth map to a rich-resource based model using only Low-Resolution (LR) single-image inputs.

\begin{figure}[t]
  \centering
  \includegraphics[width = 0.7 \linewidth]{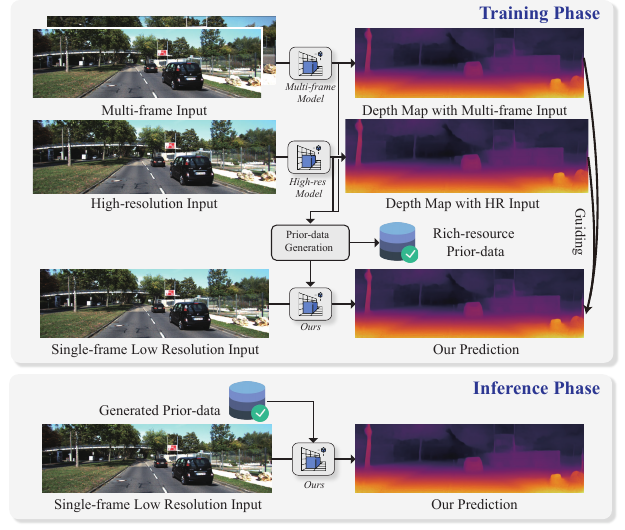}
  \caption{\textbf{Our main motivation.} In self-supervised monocular depth estimation, models using rich-resource inputs generally achieve better performance. We aim to extract prior data from rich-resource inputs during offline training, using it to enhance models with single images.
  }
  \label{Fig:motivation}
\end{figure}

To address this issue, we introduce a new self-supervised method for monocular depth estimation. The proposed method leverages features extracted from rich-resource inputs as prior information, allowing the accurate depth estimation using only LR single-image inputs during inference, as shown in Fig.~\ref{Fig:motivation}.

To be specific, our approach pivots on the idea that while rich-resource inputs (like future frames) are challenging to obtain in application, they are accessible during the training phase. This availability allows for their utilization in guiding a LR single-image input model to enhance performance. Our methodology improves model performance with rich-resource guidance in two fundamental aspects.
\textbf{Firstly}, we consider the features extracted from inputs with rich resources as a form of prior information. To achieve this, we utilize a collected generalized dataset with rich resources as a reference dataset. When estimating the depth for a LR single-image input, we initially search for similar pixels from the reference dataset. These pixels, which represent objects with similar geometric relationships, can offer valuable prior information for the model. With this prior information, the single-input model can perform similarly to rich-resource models.
\textbf{Secondly}, we investigate the intrinsic consistency present in rich-resource model predictions. We observe that rich-resource models exhibit superior geometry consistency, particularly around object edges, compared to their LR single-image counterparts. Leveraging this consistency information enhances the performance of the LR single-image model, especially in areas where depth estimation is traditionally challenging. 

In addition, we propose a feature selection algorithm to reduce the computation burden of searching reference features during inference. This algorithm effectively reduces the search space for the appropriate prior features while maintaining the same performance. Experimental results demonstrate that our method can achieve similar performance to rich-resource models when only LR single-image inputs are available. This increases the feasibility of using the depth estimation method in real-world applications. Our contributions can be summarized in four folds:

\begin{itemize}
    \item We propose a new approach for self-supervised monocular depth estimation that reduces the necessity for rich-resource, such as high-resolution, multi-frame and future frame data, while still achieving superior performance compared to models that depend on such inputs.

    \item We propose incorporating a Prior Depth Fusion Module to effectively utilize the prior information obtained from rich-resource inputs.

    \item We propose the Rich-resource Guided Loss by considering the depth prediction from rich-resource inputs as a pseudo label. This approach harnesses the consistency embedded in the pseudo label to enhance the quality of the LR single-image model.

    \item We introduce an attention-guided feature selection algorithm to reduce the computation of searching for prior depth information during inference. With this improvement, our model can achieve state-of-the-art performance while maintaining high processing speed with only LR single-image inputs.
\end{itemize}

\section{Related Work}
\subsection{Supervised Monocular Depth Estimation}
Supervised monocular depth estimation remains a core focus in computer vision, particularly for its applications in areas like autonomous vehicles and robotic navigation. This method relies on single images to infer depth maps, where each pixel value corresponds to the distance from the camera lens. In this domain, the supervised approach~\cite{zhou2022self,hui2022rm,xu2022self,swami2022you,agarwal2022depthformer,swami2022you,dijk2019neural,zhao2022monovit,zhang2023lite} necessitates ground truth depth data for training, presenting both opportunities and challenges.

The groundwork in this field was proposed by Eigen and colleagues~\cite{eigen_split}, who innovatively utilized a deep learning model for depth prediction under supervised conditions. Their model's architecture featured a dual network setup, one for coarser depth perception and another for capturing fine-grained depth details. Following this pioneering work, several researchers have contributed to refining this approach. For instance, Li \textit{et al.}~\cite{li2015depth} introduced the use of conditional random fields to enhance depth predictions, providing a new dimension to the estimation process.

Further explorations in geometry-based methods were conducted by Qi \textit{et al.}~\cite{qi2018geonet}, who proposed separate networks for estimating depth and surface normals from images. Ummenhofer \textit{et al.}~\cite{ummenhofer2017demon} contributed significantly with a network that predicts depth maps using structure from motion techniques. These advancements showcase a growing sophistication in the field. 
However, relying on extensive ground truth data, it is usually acquired through specialized equipment like LiDAR, limits the scalability and cost-effectiveness of these methods, and presenting an ongoing challenge for widespread application.

\subsection{Self-supervised Monocular Depth Estimation}
To mitigate the challenges associated with labeled data in monocular depth estimation, Garg \textit{et al.}~\cite{garg2016unsupervised} pioneered a self-supervised learning methodology. This approach used stereo images during training, aiming to minimize the disparity between synthesized and real images, marking a significant shift from traditional supervised methods.

Building upon this, Zhou \textit{et al.}~\cite{garg2016unsupervised} introduced a novel technique that estimated both the depth map and camera pose using single-camera video sequences. This method enabled the creation of artificial frames, facilitating the computation of disparities with real frames. However, this approach faced challenges such as occlusion and the presence of moving objects, which impacted the accuracy of depth estimation.
Addressing these issues, Godard \textit{et al.}~\cite{monodepth2} introduced a new minimum loss approach, exploiting the complementary nature of occlusions in adjacent frames. This allowed the model to selectively compute losses in visible areas, enhancing the accuracy of depth predictions. To address the moving object problem, they devised a strategy to ignore loss values from such objects, further refining the depth estimation process.

Subsequent research in this area has seen a variety of innovative approaches~\cite{ranjan2019competitive, zou2018df, han2024asymmetric, ranftl2020towards,lee2019big,gordon2019depth,RMSFM,pillai2019superdepth,poggi2018learning}. Masoumian \textit{et al.}~\cite{masoumian2021absolute} developed a multi-scale monocular depth estimation method using graph convolutional networks, offering a new perspective in this field. Guizilini \textit{et al.}~\cite{guizilini20203d} proposed a 3D packing network, introducing a novel architecture in depth estimation. Watson \textit{et al.}~\cite{watson2021temporal} incorporated cost volumes to build a multi-frame model, demonstrating significant improvements in depth accuracy. Furthermore, Zhou and colleagues~\cite{zhou_diffnet} explored the integration of semantic information to enhance depth estimation, indicating the potential of combining different data types for improved results.

Despite these advancements, most of the best performance models in this area rely on rich-resource data as input, which limits their application in scenarios where capturing rich-resource data is difficult. This has motivated us to develop a depth estimator that utilizes only low-resolution single-image data but still produces highly accurate depth maps.

\section{Method}
In this section, we will provide information on the proposed Rich-resource Prior Depth estimator (RPrDepth). In Sec.~\ref{sec:model}, we will explain the pipeline of the proposed method and how it can be trained end-to-end. In Sec.~\ref{sec:fusion}, we will introduce the core module of our method, the Prior Depth Fusion Module. Then in Sec.~\ref{sec:loss}, we will provide detailed information about the Rich-resource Guided Loss and how it guides the optimization of the model prediction. Finally, in Sec.~\ref{sec:selection}, we will discuss the attention-guided feature selection algorithm for reducing computation in the feature searching during the inference phase.

\begin{figure*}
  \centering
  \includegraphics[width = 0.99 \linewidth]{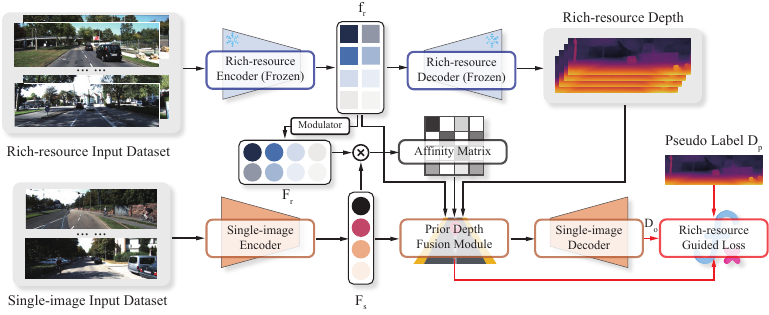}
  \caption{\textbf{Illustration of the Training Phase of Our Pipeline.} Our pipeline comprises two branches: rich-resource and LR single-image. The former generates precise depth maps and features from rich-resource images, while the latter leverages these features to achieve comparable performance.
  }
  \label{Fig:network}
\end{figure*} 

\subsection{Rich-resource Prior Depth Estimator}
\label{sec:model}
Rich-resource inputs, such as high-resolution images, multi-frame inputs and future frames, are valuable for the depth estimation task. They provide more information compared to single-frame low-resolution images, which we refer to as LR single-image inputs in this paper. However, in real-world scenarios, these rich-resource inputs are not always available, limiting the application of methods that rely on them. To address this issue, we propose a LR single-image depth estimator named Rich-resource Prior Depth estimator that bridges the performance gap between the two types of input data.

From a general perspective, LR single-image input cannot achieve the same performance as rich-resource inputs, as they lack the critical information encoded in the rich-resource inputs. For instance, when using multi-frame images as inputs, the model leverages the disparity between adjacent frames. However, LR single-image inputs do not possess this information and therefore cannot directly achieve similar performance. In this paper, we propose searching for the necessary information from the archived rich-resource inputs to bridge the information gap between the two types of inputs. To be specific, we prepare a sub-dataset called \textit{ref-dataset} which consists of rich-resource data with a wide range of variations. When we receive a LR single-image image, we search for similar feature pixels from the \textit{ref-dataset}. These pixels come from similar objects with similar geometry relationships, but they contain rich-resource data. We can use this data to fill in the missing information in the LR single-image.

Fig.~\ref{Fig:network} provides an overview of the training pipeline for our Rich-resource Prior Depth estimator. The pipeline consists of two branches: the upper branch is propsoed for rich-resource guidance, while the lower branch represents the LR single-image model. Our method is designed to be general-purpose, allowing for the use of a multi-frame model such as~\cite{manydepth} or a high-resolution based model~\cite{monodepth2} for the rich-resource guidance. During the training phase, the rich-resource model remains fixed without gradient computation.

When training the model, we begin by selecting two distinct batches $I_r, I_s$ from the \textit{ref-dataset} and the LR single-image training dataset. Next, we calculate the image features using the rich-resource encoder ($\text{Encoder}_r$) and the LR single-image encoder ($\text{Encoder}_s$), respectively:
\begin{equation}
        f_r = \text{Encoder}_\text{r}(I_r);
        F_s = \text{Encoder}_\text{s}(I_s).
\end{equation}
After that, we use a convolution module to adjust the dimension of $f_r$ to match that of $F_s$:
\begin{equation}
        F_r = \text{Conv}_\text{m}(f_r).
\end{equation}
To identify the most similar pixel in the reference dataset, we calculate the affinity between the target pixels and the reference pixels:
\begin{equation}
        \mathcal{A} = \text{Softmax}(F_s \otimes F_r).
\end{equation}
Then, we generate the rich-resource depth map $D_r$ by passing $f_r$ into the rich-resource depth decoder:
\begin{equation}
        D_r = \text{Decoder}_\text{r}(f_r).
\end{equation}
To efficiently extract and fuse the critical prior information encoded in the ref-features, we propose the Prior Depth Fusion Module. This module takes $\mathcal{A}$, $f_r$, and $D_r$ as input and produces a prior information-rich feature $F_o$ as output. Finally, the values of $F_o$ are passed to the LR single-image decoder to generate depth predictions $D_o$:
\begin{equation}
        D_o = \text{Decoder}_\text{s}(F_o).
\end{equation}
Finally, these predictions are used to construct the Rich-resource Guided Loss function. The entire pipeline is trained in an end-to-end manner using this loss function. Notably, the mentioned pipeline, which accepts both rich-resource and LR single-image inputs, is only used during the training phase. In the inference phase, the pipeline is adjusted to  accept only LR single-image inputs, as explained in Sec.~\ref{sec:selection}.

\subsection{Prior Depth Fusion Module}
\label{sec:fusion}
In our Prior Depth Fusion Module, we have designed two types of fusion procedures to effectively extract and fuse features from the ref-dataset. These procedures are the pixel-wise fusion and the depth-hint fusion. The pixel-wise fusion is responsible for completing the missing prior information in LR single-image data using the corresponding rich-resource data as a reference. To efficiently identify the most similar pixel, we add an auxiliary loss to guide the search process. On the other hand, the depth-hint fusion aggregates the prior information from the entire ref-dataset in an attention manner, without any explicit guidance.

\begin{figure*}[t]
  \centering
  \includegraphics[width = 0.4 \linewidth]{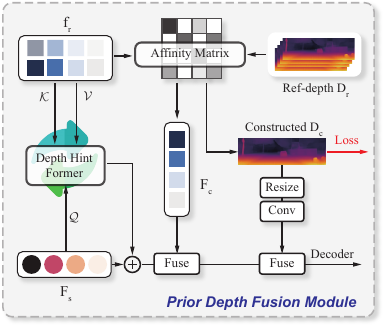}
  \caption{\textbf{Illustration of the Prior Depth Fusion Module.} 
  }
  \label{Fig:pdfm}
\end{figure*}

Fig.~\ref{Fig:pdfm} shows an illustration of the Prior Depth Fusion Module. In this module, we first use a transformer module to extract and fuse the depth-hint prior information from the reference features. In this transformer, we consider the reference feature $f_r$ as the key $K$ and value $V$, and the target feature $F_s$ as the query $Q$.
Then we employ the multi-head attention to fuse the depth-hint information and produce the output feature $F_d$:
\begin{equation}
        F_d = \text{MHA}(\text{Q}, \text{K}, \text{V}).
\end{equation}
Next, we need to address the pixel-wise prior information. In the pipeline, we have calculated the affinity between the target and reference pixels. Using the affinity, $f_r$ and $D_r$, we can construct a pixel-wise constructed reference feature map $F_c$ and constructed reference depth $D_c$:
\begin{equation}
        F_c = \mathcal{A} \times f_r;
        D_c = \mathcal{A} \times D_r. 
\end{equation}
Afterwards, we combine $F_c$ and $F_s$ and apply a convolution module to compress the feature map to its original number of channels. In addition to the reference features, we also take into account the output depth of the rich-resource model as valuable prior information. Consequently, we regard the prior depth map $D_c$ as a reference and merge it with the feature map, similar to the reference feature, as shown in the figure. Furthermore, the constructed depth map is utilized to formulate an auxiliary loss. In this process, we minimize the discrepancy between the constructed depth map and the prediction of the high-resource model. This loss function can aid in guiding the optimization of the affinity matrix.

Notably, the Prior Depth Fusion Module involves calculating the attention matrix between the target batch and the reference batch, an operation with a space complexity of $O(MN)$, where M and N are the pixel numbers of the target and reference batches, respectively. To enhance the representation of the reference data, we should use a relatively large size for the reference batch. However, this will result in a significant memory burden. To address this limitation and make the pipeline easier to train,  we randomly sample features from the whole reference dataset offline (1\% pixels from 2000 images). This is based on the observation that adjacent pixels have similar geometric information. Therefore, when we randomly sample from a large batch, the selected pixels can provide more contextual conditions than selecting all pixels from a smaller batch.

\begin{figure}[t]
  \centering
  \includegraphics[width = 0.6 \linewidth]{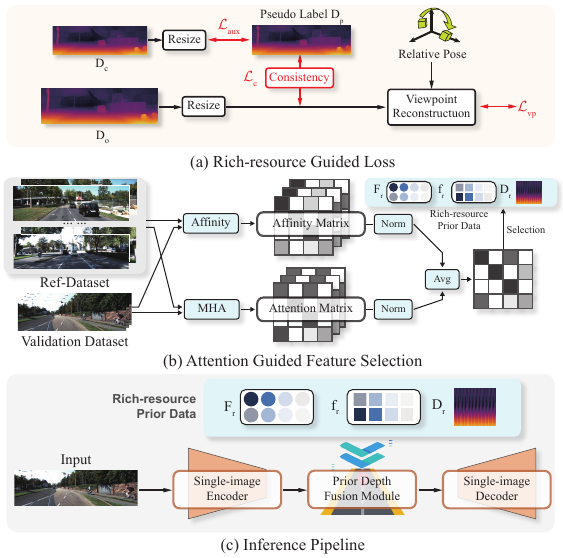
  }
  \caption{\textbf{Illustration of the Loss and Inference Pipeline.} 
  (a) Illustration of the Rich-resource guided loss. (b) Illustration of Attention Guided Feature Selection. (c) The Inference Pipeline of RPrDepth.
  }
  \label{Fig:modules}
\end{figure} 

\subsection{Rich-resource Guided Loss}
\label{sec:loss}
To enhance the performance of the proposed pipeline and the Prior Depth Fusion Module, we recommend incorporating the Rich-resource Guided Loss, as shown in Fig.~\ref{Fig:modules} (a). This loss function effectively utilizes guidance from both rich-resource inputs and the rich-resource model predictions to optimize the LR single-image model. The proposed loss function consists of two parts: the viewpoint reconstruction loss guided by rich-resource inputs and the consistency loss guided by the predictions of the rich-resource model.


Most of the self-supervised monocular depth estimation methods use the viewpoint reconstruction loss to guide the model optimization. Following the previous method~\cite{monodepth2}, we also use the viewpoint reconstruction as the main loss function. As we have rich-resource inputs available during the training phase, we choose to reconstruct the new viewpoint images from these inputs. These inputs contain more detailed information and can provide more accurate guidance for the model. To bridge the resolution gap between our prediction and the rich-resource inputs, we upsample the prediction with cubic interpolation algorithm to match the size of the target image, and the final reconstruction loss is defined as:
\begin{equation}
    \mathcal{L}_{\text{vp}} = l_{vp}\left( \text{Resize}(D_o), I_r \right)
\end{equation}

In addition to the reconstruction loss, we also leverage the pseudo labels generated by the rich-resource model to optimize the LR single-image model. rich-resource models typically produce depth maps with greater detail accuracy, particularly in the edge areas, compared to ordinary LR single-image depth estimators. Therefore, it is meaningful to utilize the advantages of rich-resource predictions to instruct the LR single-image model. However, directly using the pseudo label as the target and minimizing the difference between the predictions and the pseudo labels is not a desirable approach. Since the models are trained in a self-supervised manner, their predictions represent relative disparity rather than accurate depth values. As a result, different models may have variations in scale in their predictions. Hence, we choose to minimize the gradients along the x and y axes between the prediction and the rich-resource predictions.

Specifically, we start by calculating the gradient on the x and y-axis of the output depth map $D_o$ and the pseudo label generated offline $D_p$. It's important to note that despite both $D_p$ and $D_r$ being depth maps generated by the rich-resource model, they're derived from different batches. Specifically, $D_r$ comes from the reference batch, while $D_p$ is from the training batch. Next, we normalize the two gradient maps and add up the x and y-gradients.
\begin{equation}
\begin{aligned}
       \tilde{G}_{x,y}(D_o) &= \text{Norm}(\nabla_x D_o) + \text{Norm}(\nabla_y D_o)  \\
\tilde{G}_{x,y}(D_p) &= \text{Norm}(\nabla_x D_p) + \text{Norm}(\nabla_y D_p).
\end{aligned}
\end{equation}
We use an L1 loss to minimize the gradient difference:
\begin{equation}
    \mathcal{L}_{\text{c}} = \| \tilde{G}_{x,y}(D_o) - \tilde{G}_{x,y}(D_p) \|_1.
\end{equation}
Additionally, as mentioned in the previous section, we use an auxiliary loss $L_\text{aux}$ to guide the optimization of the affinity matrix. To achieve this, we up-sample the constructed depth-map $D_c$ to the same size as the pseudo label $D_p$. We then minimize the difference between them directly. Since $D_c$ is constructed directly from the pixels of the high-resource model prediction, it must have the same scale factor as $D_p$. Therefore, we can simply minimize their difference rather than the gradient:
\begin{equation}
    \mathcal{L}_{\text{aux}} = \| D_p - \text{Resize}(D_c)\|_1.
\end{equation}
The final loss is determined by the combination of the reconstruction loss, the consistency loss and the auxiliary loss:
\begin{equation}
    \mathcal{L} = \alpha  \mathcal{L}_{\text{vp}} + \beta  \mathcal{L}_{\text{c}} + \mathcal{L}_{\text{aux}},
\end{equation}
where $\alpha,\beta$ are the balance ratios. 


\subsection{Attention Guided Feature Selection}
\label{sec:selection}
As mentioned in the previous sections, the number of the reference dataset is crucial. The reference dataset should have sufficient variety to encompass all possible conditions that may be found in the target image. However, if the size of this reference dataset is too large, it may result in a significant computational burden when searching for the reference pixels.
To overcome this limitation, we propose a new solution that involves using a subset of the reference dataset instead of the entire dataset during the inference phase. This subset is selected to be the most representative of the reference dataset. To achieve this, we introduce the attention-guided feature selection algorithm, as shown in Fig.~\ref{Fig:modules} (b). The proposed algorithm selects the features from the reference dataset in an attention-based manner.

In the depth-hint fusion procedure, the reference features are used with multi-head attention, while in the pixel-wise fusion procedure, the features are incorporated with a learnable affinity matrix. By leveraging these two weight matrices, we can determine which pixels are more important for the target image. We then aggregate the weight matrices across the entire validation dataset and calculate the average weight matrix for each pixel in the reference dataset:
\begin{equation}
    W_{\text{avg}} = \frac{1}{N} \sum_{i=1}^{N} (\mathcal{A}_i + \mathcal{A}_{\text{MHA},i})
\end{equation}
Finally, we sort the pixels in the matrix and select the ones with the highest weight to represent the entire reference dataset.

We store the pixels with the highest weight value. These pixels serve as the depth-prior data, which remains unchanged during the inference phase. Once the depth-prior data is generated, we replace it with the original rich-resource model in our pipeline and fine-tune the LR single-image model for a few epochs. Surprisingly, we find that the performance of the LR single-image model does not decrease due to the decrease in computation, but actually outperforms the original model. We attribute this to the fact that the selected pixels are more concise and meaningful, contributing to the improved performance. By eliminating other irrelevant pixels that may cause interference, the model can better learn to utilize this prior information.

The final inference pipeline is shown in Fig.~\ref{Fig:modules}(c). In comparison to the original ref-dataset based pipeline, the compressed prior data is only less than 1\% of the original size (from 2,560,000 to 25,000 pixels), significantly reducing the computational load. 

\section{Experiment}
For all the training and evaluation processes, we utilize one work station with a single V100 GPU. To demonstrate the enhancements, we integrate these advancements alongside a recent, highly efficient baseline known as DIFFNet~\cite{zhou_diffnet}, inspired by the HR-Net networks~\cite{sun2019high,hrdepth}. For the high-resource guidance, we have opted for ManyDepth~\cite{manydepth}  as our choice. ManyDepth is a well-known baseline model that utilizes multi-frame images as input. During the training of our model, we specifically chose the HR version of ManyDepth to provide the feature prior and loss guidance. This model accepts multi-frame, high-resolution images along with future frames as input.

\subsection{Comparison on KITTI}
The KITTI dataset stands out as a widely utilized benchmark in the field of computer vision.
It's also highly regarded as a benchmark in the area of self-supervised monocular depth estimation.
Our approach utilizes the data partitioning strategy mentioned in~\cite{eigen_split} as a foundation for our models, and we follow the preprocessing steps outlined in~\cite{zhou2017unsupervised} to eliminate static frames. During the training phase, we randomly select 2,000 triplets from the training dataset as the reference dataset, and use the remaining 37, 810 triplets as the training data. When running the feature selection procedure, we use a separated validation set as the target, ensuring that it has no overlap with the test set.

\begin{table*}[t]
\caption{\textbf{The SOTA comparison on KITTI Eigen Split~\cite{eigen_split}}. 
We evaluate our methods against established models on this benchmark, using three self-supervision techniques: ``\textbf{M}" for monocular videos, ``\textbf{S}" for stereo images, and ``\textbf{MS}" for both.
The best and second-best results are marked in \textbf{bold} and \underline{underline}, respectively.
}
\label{Table:sota}
\small
\centering
  \resizebox{\textwidth}{!}
  {
\setlength{\tabcolsep}{1mm}{
\begin{tabular}{l|c|c|c|c|c|c|c|c|c|c}
\bottomrule
{{Method}} &  TestFrames &
{{Resolution}} & {{Trian}}  &                {\cellcolor{red!30}Abs Rel} & {\cellcolor{red!30}Sq Rel} & {\cellcolor{red!30}RMSE}  & {\cellcolor{red!30}RMSE log} & {\cellcolor{cyan!20}$\delta < 1.25$} & {\cellcolor{cyan!20}$\delta < 1.25^{2}$} & \cellcolor{cyan!20} $\delta < 1.25^3$ \\ \hline

\hline
\hline
{Monodepth2~\cite{monodepth2}}	& 1&{$640 \times 192$}	&	{M}	&	{0.115}	&	{0.903}	&	{4.863}	&	{0.193}	&	{0.877}	&	{0.959}	&	\multicolumn{1}{c}{0.981} \\
{PackNet-SfM~\cite{guizilini20203d}}	&	1&{$640 \times 192$}	&	{M}	&	{0.111}	&	{{0.785}}	&	{{4.601}}	&	{0.189}	&	{0.878}	&	{0.960}	&	\multicolumn{1}{c}{0.982} \\
{HR-Depth~\cite{hrdepth}}	&	1&{$640 \times 192$}	&	{M}	&	{0.109}	&	{0.792}	&	{4.632}	&	{{0.185}}	&	{0.884}	&	{{0.962}}	&	\multicolumn{1}{c}{{0.983}} \\
{DIFFNet~\cite{zhou_diffnet}}	&	1&{$640 \times 192$}	&	{M}	&	{0.102}	&	{0.764}	&	{4.483}	&	{0.180}	&	{0.896}	&	{0.965}	&	\multicolumn{1}{c}{0.983} \\
{BRNet~\cite{brnet}}	&	1&{$640 \times 192$}	&	{M}	&	{{0.105}}	&	{\underline {0.698}}	&	{{4.462}}	&	{{0.179}}	&	{{0.890}}	&	{{0.965}}	&	\multicolumn{1}{c}{\underline {0.984}} \\
{MonoFormer~\cite{monoformer}}	&	1&{$640 \times 192$}	&	{M}	&	{{0.108}}	&	{{0.806}}	&	{{4.594}}	&	{{0.184}}	&	{{0.884}}	&	{{0.963}}	&	\multicolumn{1}{c}{{0.983}} \\
{Lite-Mono~\cite{litemono}}	&	1&{$640 \times 192$}	&	{M}	&	{{0.107}}	&	{{0.765}}	&	{{4.561}}	&	{{0.183}}	&	{{0.886}}	&	{{0.963}}	&	\multicolumn{1}{c}{{0.983}} \\
{Wang \textit{et al.}}~\cite{wang2020selfsupervised}	&	2 (-1, 0) &{$640 \times 192$}	&	{M}	&	{{0.106}}	&	{{0.799}}	&	{{4.662}}	&	{{0.187}}	&	{{0.889}}	&	{{0.961}}	&	\multicolumn{1}{c}{{0.982}} \\
{ManyDepth~\cite{manydepth}}	&	2 (-1, 0) &{$640 \times 192$}	&	{M}	&	{\underline {0.098}}	&	{{0.770}}	&	{\underline {4.459}}	&	{\underline {0.176}}	&	{\textbf{0.900}}	&	{\underline {0.965}}	&	\multicolumn{1}{c}{{0.983}} \\
\rowcolor{green!10}{RPrDepth~(ours)}	&	1&{$640 \times 192$}	&	{M}	&	\textbf{{0.097}}	&	\textbf{{0.658}}	&	\textbf{{4.279}}	&	\textbf{{0.169}}	&	\textbf{{0.900}}	&	\textbf{{0.967}}	&	\multicolumn{1}{c}{\textbf{0.985}} \\
\hline
{Monodepth2~\cite{monodepth2}}	&	1&{$640 \times 192$}	&	{S}	&	{{0.109}}	&	{{0.873}}	&	{{4.960}}	&	{{0.209}}	&	{{0.864}}	&	{{0.948}}	&	\multicolumn{1}{c}{0.975} \\
{BRNet~\cite{brnet}}	&	1&{$640 \times 192$}	&	{S}	&	{\underline{0.103}}	&	{\underline{0.792}}	&	{\underline{4.716}}	&	{\underline{0.197}}	&	{\underline{0.876}}	&	{\underline{0.954}}	&	\multicolumn{1}{c}{\underline{0.978}} \\
\rowcolor{green!10}{RPrDepth~(ours)}	&	1&{$640 \times 192$}	&	{S}	&	{\textbf{0.098}}	&	{\textbf{0.716}}	&	{\textbf{4.538}}	&	{\textbf{0.185}}	&	{\textbf{0.885}}	&	{\textbf{0.960}}	&	\multicolumn{1}{c}{\textbf{0.980}} \\
\hline
{HR-Depth~\cite{hrdepth}}	&	1&{$640 \times 192$}	&	{MS}	&	{0.107}	&	{{0.785}}	&	{{4.612}}	&	{{0.185}}	&	{{0.887}}	&	{{0.962}}	&	\multicolumn{1}{c}{{0.982}} \\
{DIFFNet~\cite{zhou_diffnet}}	&	1&{$640 \times 192$}	&	{MS}	&	{0.101}	&	{0.749}	&	\underline{4.445}	&	\underline{0.179}	&	\underline{0.898}	&	\underline{0.965}	&	\multicolumn{1}{c}{\underline{0.983}} \\
{BRNet~\cite{brnet})}	&	1&{$640 \times 192$}	&	{MS}	&	{\underline{0.099}}	&	{\underline{0.685}}	&	{{4.453}}	&	{{0.183}}	&	{{0.885}}	&	{{0.962}}	&	\multicolumn{1}{c}{\underline{0.983}} \\
\rowcolor{green!10}{RPrDepth~(ours)}	&	1&{$640 \times 192$}	&	{MS}	&	{\textbf{0.095}}	&	{\textbf{0.638}}	&	{\textbf{4.232}}	&	{\textbf{0.169}}	&	{\textbf{0.902}}	&	{\textbf{0.970}}	&	\multicolumn{1}{c}{\textbf{0.985}} \\
\hline
\hline
{PackNet-SfM~\cite{guizilini20203d}}	&	1&{$1280 \times 384$}	&	{M}	&	{0.107}	&	{0.802}	&	{4.538}	&	{0.186}	&	{{0.889}}	&	{0.962}	&	\multicolumn{1}{c}{0.981} \\
{HR-Depth~\cite{hrdepth}}	&	1&{$1024 \times 320$}	&	{M}	&	{{0.106}}	&	{0.755}	&	{4.472}	&	{{0.181}}	&	{{0.892}}	&	{{0.966}}	&	\multicolumn{1}{c}{{0.984}} \\
{DIFFNet~\cite{zhou_diffnet}}	&	1&{$1024 \times 320$}	&	{M}	&	{0.097}	&	{0.722}	&	{{4.435}}	&	{{0.174}}	&	{{0.907}}	&	\underline{0.967}	&	\multicolumn{1}{c}{{0.984}} \\
{BRNet~\cite{brnet}}	&	1&{$1024 \times 320$}	&	{M}	&	{{0.103}}	&	{\underline{0.684}}	&	{{4.385}}	&	{{0.175}}	&	{{0.889}}	&	{{0.965}}	&	\multicolumn{1}{c}{\underline {0.985}} \\
{Lite-Mono~\cite{litemono}}	&	1&{$1024 \times 320$}	&	{M}	&	{{0.102}}	&	{{0.746}}	&	{{4.444}}	&	{{0.179}}	&	{{0.896}}	&	{{0.965}}	&	\multicolumn{1}{c}{{0.983}} \\
{Wang \textit{et al.}~\cite{wang2020selfsupervised}}	&	2 (-1, 0) &{$1024 \times 320$}	&	{M}	&	{{0.106}}	&	{{0.773}}	&	{{4.491}}	&	{{0.185}}	&	{{0.890}}	&	{{0.962}}	&	\multicolumn{1}{c}{{0.982}} \\
{ManyDepth-HR~\cite{manydepth}}	&	2 (-1, 0) &{$1024 \times 320$}	&	{M}	&	{\underline {0.093}}	&	{{0.715}}	&	{\underline {4.245}}	&	{\underline {0.172}}	&	{\underline {0.909}}	&	{{0.966}}	&	\multicolumn{1}{c}{{0.983}} \\
\rowcolor{green!10}{RPrDepth~(ours)}	&	1& {$1024 \times 320$}	&	{M}	&	{\textbf{0.091}}	&	{\textbf{0.612}}	&	{\textbf{4.098}}	&	{\textbf{0.162}}	&	{\textbf{0.910}}	&	{\textbf{0.971}}	&	\multicolumn{1}{c}{\textbf{0.986}} \\
\hline
{Monodepth2~\cite{monodepth2}}	&	1& {$1024 \times 320$}	&	{S}	&	{{0.107}}	&	{{0.849}}	&	{{4.764}}	&	{{0.201}}	&	{{0.874}}	&	{{0.953}}	&	\multicolumn{1}{c}{{0.977}} \\
{BRNet~\cite{brnet}}	&	1&{$1024 \times 320$}	&	{S}	&	{\underline{0.097}}	&	{\underline{0.729}}	&	{\underline{4.510}}	&	{\underline{0.191}}	&	{\underline{0.886}}	&	{\underline {0.958}}	&	\multicolumn{1}{c}{\textbf{0.979}} \\
\rowcolor{green!10}{RPrDepth~(ours)}	&	1&{$1024 \times 320$}	&	{S}	&	{\textbf{0.091}}	&	{\textbf{0.689}}	&	{\textbf{4.412}}	&	{\textbf{0.185}}	&	{\textbf{0.892}}	&	{\textbf{0.959}}	&	\multicolumn{1}{c}{\textbf{0.979}} \\
\hline
{HR-Depth~\cite{hrdepth}}	&	1&{$1024 \times 320$}	&	{MS}	&	{{0.101}}	&	{{0.716}}	&	{{4.395}}	&	{{0.179}}	&	{{0.899}}	&	{{0.966}}	&	\multicolumn{1}{c}{{0.983}} \\
{BRNet~\cite{brnet}}	&	1&{$1024 \times 320$}	&	{MS}	&	{{0.097}}	&	{{0.677}}	&	{{4.378}}	&	{{0.179}}	&	{{0.888}}	&	{{0.965}}	&	\multicolumn{1}{c}{\underline{0.984}} \\
{DIFFNet~\cite{zhou_diffnet}}	&	1&{$1024 \times 320$}	&	{MS}	&	{\underline{0.094}}	&	{\underline{0.678}}	&	{\underline{4.250}}	&	\underline{{0.172}}	&	{\underline{0.911}}	&	{\underline{0.968}}	&	\multicolumn{1}{c}{\underline{0.984}} \\
\rowcolor{green!10}{RPrDepth~(ours)}	&	1&{$1024 \times 320$}	&	{MS}	&	{\textbf{0.089}}	&	{\textbf{0.613}}	&	{\textbf{4.120}}	&	{\textbf{0.159}}	&	{\textbf{0.913}}	&	{\textbf{0.970}}	&	\multicolumn{1}{c}{\textbf{0.985}} \\
\hline
\end{tabular}}}
\end{table*}

\noindent\textbf{SOTA comparison} Table~\ref{Table:sota} presents our RPrDepth's assessment on the Eigen split~\cite{eigen_split}, categorizing results by low and high resolutions. We utilize seven metrics for comparison, with $Abs Rel$, $Sq Rel$, $RMSE$, $RMSE log$ as error metrics where lower scores indicate better performance. Conversely, $\delta$ measures the deviation from actual depth values, with $\delta < 1.25$, $\delta < 1.25^2$, $\delta < 1.25^3$ being accuracy metrics where higher scores are favorable. Our RPrDepth tops all categories in terms of supervision types and resolutions.

As shown in this table, our model with LR single-image input outperforms our baseline model DIFFNet, and even outperforms the guiding model ManyDepth-HR, which is based on multi-frame high-resolution inputs. 
Notably, the performance of ManyDepth in this table is without future frames, because future frames are not available during inference. 

\begin{figure*}[t]
\centering
\resizebox{0.8\textwidth}{!}
  {
  \includegraphics[width = 15cm]{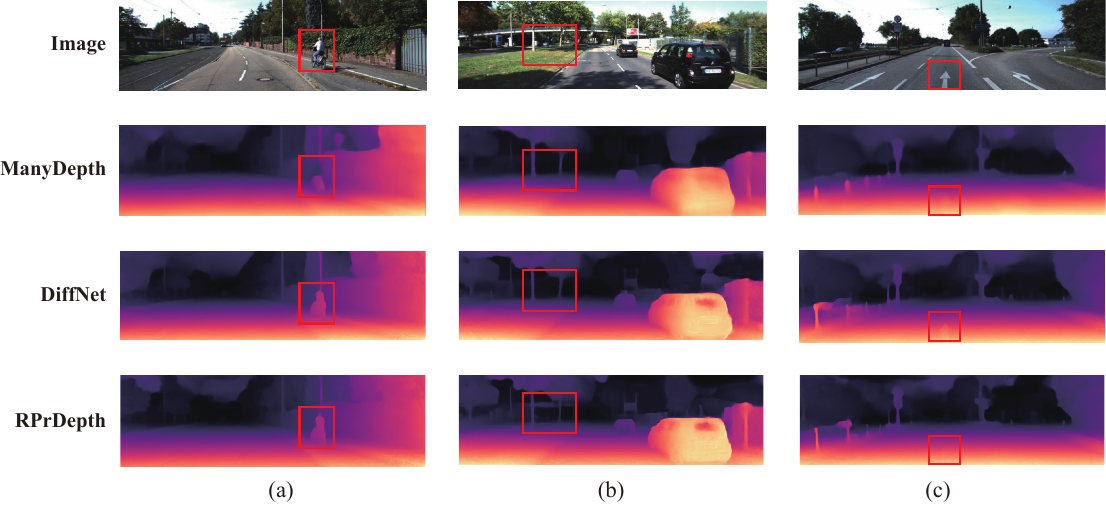}}
  \caption{\textbf{Qualitative results on the KITTI Eigen split test set.} Our RPrDepth can correct the errors of both LR single-image models and rich-resource based models.}
  \label{Fig:qualitive}
\end{figure*} 

\noindent\textbf{Qualitative Results} 
Fig.~\ref{Fig:qualitive} shows a comparison between our model, DIFFNet, and the guiding model ManyDepth. In comparison to models with rich-resource input, our model performs better on moving objects, as demonstrated in Fig.~\ref{Fig:qualitive} (a). This is because multi-frame based methods are not well-suited for moving objects~\cite{manydepth}. However, our model can identify relevant information for moving objects and correct the issue. Additionally, compared to other single image models, our model can address incorrect depth predictions caused by texture, as seen in Fig.~\ref{Fig:qualitive} (c) with the arrow on the road. Ordinary LR single-image models struggle to distinguish texture, but our model leverages prior information from rich references to solve this problem.

\subsection{Comparison on Make3D and Cityscapes}
\noindent\textbf{Make3D} dataset is composed of both single-camera RGB images and their related depth maps. It lacks stereo images and monocular sequences, rendering it unsuitable for training self-supervised monocular depth estimation models. However, it is commonly used as a test set to assess the performance of networks on a varied dataset. In our study, we evaluated our models against other notable research in this area. The results, as detailed in Table~\ref{Table:make3d}, reveal that our models surpass all competing methods, indicating their robustness in adapting to novel environments. Utilizing monocular training with a resolution of $640 \times 192$, our approach records $0.288$ in $Abs Rel$ and $6.532$ in $RMSE$, markedly surpassing other leading models in performance.

\noindent\textbf{Citiscapes} dataset stands as a key resource in the field of semantic segmentation, particularly for autonomous driving applications. It encompasses a collection of stereo video sequences, which are instrumental for training self-supervised depth estimation models. Adhering to the approach outlined in~\cite{watson2021temporal}, we conducted training and evaluation of our RPrDepth model using the Cityscape dataset. The outcomes, as presented in Table~\ref{Table:city}, demonstrate that RPrDepth remarkably exceeds the performance of numerous advanced models.

\begin{table}[t!]
\caption{\color{black}\textbf{Make3D results with monocular training and $640\times192$ inputs.}
}
\label{Table:make3d}
\footnotesize
\centering
\resizebox{0.8\textwidth}{!}{
\setlength{\tabcolsep}{4mm}{
\begin{tabular}{c|c|c|c|c}
\hline
Architecture & Abs Rel $\downarrow$ & Sq Rel $\downarrow$ & RMSE$\downarrow$ & $log_{10}$  $\downarrow$ \\
\hline
\hline
Monodepth2    & 0.322                & 3.589             & 7.414 & 0.163 \\
\hline
BRNet    & {0.302}                & {3.133}             & {7.068} & {0.156} \\
\hline
\rowcolor{gray!10} RPrDepth    & \textbf{0.288}                & \textbf{2.868}             & \textbf{6.532} & \textbf{0.145} \\
\hline
\end{tabular}
}}
\end{table}

\begin{table}[t!]
\caption{\color{black}\textbf{Cityscape results follow the settings of~\cite{watson2021temporal}.}
}
\label{Table:city}
\footnotesize
\centering
\resizebox{0.8\textwidth}{!}{
\setlength{\tabcolsep}{3mm}{
\begin{tabular}{c|c|c|c|c}
\hline
Architecture & Frames & Abs Rel $\downarrow$ &  RMSE$\downarrow$ & {$\delta < 1.25$ $\uparrow$} \\
\hline
\hline
Monodepth2~\cite{monodepth2}   & 1 & 0.129               & 6.876 & 0.849 \\
\hline
Li \textit{et al.}~\cite{li2021unsupervised}   & 1 & 0.119                & 6.980 & 0.846 \\
\hline
ManyDepth~\cite{manydepth}   & 2 (-1, 0) & {\underline {0.114}}          & \textbf{6.223} & {\underline {0.875}} \\
\hline
\rowcolor{gray!10} RPrDepth   & 1 & \textbf{0.111}                & {\underline {6.243}} & \textbf{0.890} \\
\hline
\end{tabular}
}}
\end{table}

\subsection{Ablation Study}
We performed several ablation studies on the KITTI dataset. We used the Eigen split~\cite{eigen_split} to validate the effectiveness of the proposed modules: Prior Depth Fusion (PDF) module, Attention Guided Feature Selection (AGFS), and Rich-resource Guided Loss (RGL). Specifically, +PDF indicates the baseline model with rich-resource feature prior, +AGFS module indicates that we replaced the reference dataset with the selected features, and +RGL indicates that the model was trained with the proposed new loss function. Lastly, +Full indicates the model with all the components.

As shown in Table~\ref{Table:ablation}, integrating prior information from rich resource data significantly improves the model across all metrics. After applying the feature selection algorithm, we further enhance the performance, particularly in terms of the $RMSE$ metric. Additionally, the computational burden of the search process is significantly reduced. Overall, the feature selection algorithm reduces the number of features to just 1\% of the entire reference dataset. The proposed RGL also clearly improves performance. Finally, the combination of all components achieves the best performance.


\begin{table}[t]
\caption{\textbf{Ablation study of the proposed RPrDepth.} 
}
\label{Table:ablation}
\centering
  \resizebox{0.7\textwidth}{!}
  {
\setlength{\tabcolsep}{6mm}{
\begin{tabular}{l|c|c|c}
\hline
\multicolumn{1}{c|}{Components} & \multicolumn{1}{c|}{\multirow{1}{*}{Abs Rel $\downarrow$}} & RMSE $\downarrow$ & \multirow{1}{*}{$\delta < 1.25$ $\uparrow$} \\ \hline\hline
 Baseline & 0.102 & 4.483 & 0.896 \\ \hline
 + PDF & 0.098 & 4.284 & 0.898 \\ \hline
 + AGFS & 0.098 & 4.240 & 0.898 \\ \hline
 + RGL & 0.100 & 4.321 & 0.897 \\ \hline
 + Full & 0.097 & 4.279 & 0.900 \\ \hline
\hline

\end{tabular}}}
\end{table}



\section{Conclusion}
In the field of self-supervised depth estimation, many top-performing models use rich-resource images as input, such as multi-frame images and high-resolution images. However, these rich-resource  inputs are not always available in real-world applications. Therefore, in this paper, we propose a new depth estimation model that leverages the prior information encoded in rich-resource images during the training and uses only a single image to generate the depth map during the inference phase. Specifically, we propose three key modules. The first module is the Prior Depth Fusion, which efficiently combines the prior features. The second module is the Rich-resource Guided Loss, which guides the optimization of LR single-image models. Lastly, we introduce the Attention Guided Feature Selection algorithm to enhance the searching efficiency from the reference images. We aim for our method to provide a new perspective on improving the practicality of high-performance depth estimation.

{\small
\bibliographystyle{ieee_fullname}
\bibliography{egbib}
}

\end{document}


\definecolor{citecolor}{HTML}{014A98}
\hypersetup{
    colorlinks=true,
    linkcolor=red,
    filecolor=magenta,      
    urlcolor=magenta,
    citecolor=citecolor,
}

\title{Supplementary:High-Precision Self-Supervised Monocular Depth Estimation \\
with Rich-Resource Prior} 

\maketitle

\section{Pseudo Codes}
To enhance the comprehension and implementation of the proposed method, we provide pseudo codes in pytorch-style for the main modules in our approach. Code.1 illustrates the training phase of our pipeline, while Code.2 showcases the feature selection algorithm.

\begin{center}
\begin{lstlisting}[caption={Pseudo Code for Rich-resource Prior Depth Estimator},label={lst:code1},linewidth=\linewidth]
# Main Training Loop
def train(model, ref_loader, train_loader, optimizer):
    for I_r, (I_s, D_p) in zip(ref_loader, train_loader):
        # Extract features
        f_r = Encoderr(I_r)
        F_s = Encoders(I_s)

        # Adjust dimensions
        F_r = Convm(f_r)

        # Calculate affinity
        A = F.softmax(F_s @ F_r.T, dim=-1)

        # Generate rich-resource depth estimations
        D_r = Decoderr(f_r)

        # Compute Prior Depth Fusion
        F_o, D_c = PriorDepthFusionModule(A, f_r, D_r)
        # Generate depth prediction
        D_o = Decoders(F_o)

        # Calculate loss and update weights
        # D_p is precomputed pseudo label loaded from dataset
        loss = rich_resource_guided_loss(D_o, D_c, D_p)
        optimizer.zero_grad()
        loss.backward()
        optimizer.step()
\end{lstlisting}
\end{center}

\begin{center}
\begin{lstlisting}[caption={Pseudo Code for Attention Guided Feature Selection},label={lst:code2},linewidth=\linewidth]
# Define Attention Guided Feature Selection Algorithm
def attentionGuidedFeatureSelection(val_dataset, ref_features, mha_func, affinity_func):
    # mha_func and affinity_func are the functions for calculating the multi-head attention maps and affinity maps.
    # ref_features are the features extracted from the whole reference dataset
    # Initialize average weight matrix
    W_avg = None
    N = len(val_dataset)

    for data in val_dataset:
        # Extracting features
        features = extract_features(data)
        
        # Pooling multi-head attention map into one channel
        A_mha = mha_func(features, ref_features).mean(1)
        # Apply affinity model for pixel-wise fusion
        A_affinity = affinity(features, ref_features)

        # Summing weights from both models
        A_combined = A_mha + A_affinity

        # Update the average weight matrix
        if W_avg is None:
            W_avg = A_combined
        else:
            W_avg += A_combined

    # Calculating the average
    W_avg /= N

    # Sorting pixels in the matrix
    indices = np.argsort(W_avg)[::-1]  # Reverse for descending order

    # Select top 25000 pixels with the highest weight
    selected_pixels = indices[:25000]  
    # Select top 25000 pixels which are about 1% of all the pixels in the reference dataset.

    return selected_pixels
\end{lstlisting}
\end{center}

\section{Improved Ground Truth}
The assessment technique developed by Eigen~\cite{eigen_split} for the KITTI dataset involves using LIDAR projections, but this method struggles with occlusions and moving objects - common issues in environments with moving vehicles. Addressing these challenges, a high-quality set of depth maps was introduced for KITTI, which incorporates data from five consecutive frames and manages moving objects using stereo pairs. This enhanced dataset includes 652 frames from the Eigen division, accounting for 93\% of the total test frames (697). Following the approach of a previous study~\cite{monodepth2}, we assess our methods using these frames with refined ground truth and compare the results against various notable networks.

In our evaluation, we adhere to the standard error metrics and limit the predicted depth to 80 meters, aligning with Eigen's evaluation criteria. The results, detailed in a referenced table, show that our methods, trained with three types of supervision, significantly outperform our initial baseline and surpass all existing methods.

\begin{table*}[h]
\centering
\small
\resizebox{1\textwidth}{!}{
\setlength{\tabcolsep}{2.5mm}{
\begin{tabular}{c|c|c|c|c|c|c|c|c|c}
\hline
\hline
\multirow{2}{*}{Method} & \multirow{2}{*}{Resolution} & \multirow{2}{*}{Train} & \multicolumn{4}{c}{lower is better}                               & \multicolumn{3}{c}{higher is better}                    \\ \cline{4-10} 
                        &                             &                        & Abs Rel        & Sq Rel         & RMSE           & RMSE log       & $\delta < 1.25$ & $\delta^2 < 1.25$ & $\delta^3 < 1.25$ \\ \hline
GeoNet~\cite{yin2018geonet}               & $416 \times 128$            & M                      & 0.132          & 0.994          & 5.240          & 0.193          & 0.883           & 0.953             & 0.985             \\
DDVO~\cite{wang2018learning}                   & $416 \times 128$            & M                      & 0.126          & 0.866          & 4.932          & 0.185          & 0.851           & 0.958             & 0.986             \\
EPC++~\cite{luo2019every}                & $640 \times 192$            & M                      & 0.120          & 0.789          & 4.755          & 0.177          & 0.856           & 0.961             & 0.987             \\
Monodepth2~\cite{monodepth2}             & $640 \times 192$            & M                      & { 0.090}    & { 0.545}    & { 3.942}    & { 0.137}    & { 0.914}     & { 0.983}       & { 0.995}       \\
BRNet~\cite{brnet}                & $640 \times 192$            & M                      & {\ul 0.080} & {\ul 0.409} & {\ul 3.613} & {\ul 0.124} & {\ul 0.928}  & {\ul 0.987}    & \textbf{0.997}    \\ 
RPrDepth                   & $640 \times 192$            & M                      & \textbf{0.069} & \textbf{0.322} & \textbf{3.025} & \textbf{0.108} & \textbf{0.945}  & \textbf{0.991}    & \textbf{0.997}    \\ \hline
SuperDepth+pp~\cite{pillai2019superdepth}        & $416 \times 128$            & S                      & 0.090          & 0.542          & 3.967          & 0.144          & 0.901           & 0.976             & {\ul 0.993}       \\
Monodepth2~\cite{monodepth2}           & $640 \times 192$            & S                      & { 0.085}    & { 0.537}    & { 3.868}    & { 0.139}    & { 0.912}     & { 0.979}       & { 0.993}       \\
BRNet~\cite{brnet}                & $640 \times 192$            & S                      & {\ul 0.078} & {\ul l0.448} & {\ul 3.547} & {\ul0.125} & {\ul 0.928}  & \textbf{0.985}    & {\ul 0.995}    \\ 
RPrDepth                   & $640 \times 192$            & S                      & \textbf{0.074} & \textbf{0.419} & \textbf{3.398} & \textbf{0.120} & \textbf{0.935}  & \textbf{0.985}    & \textbf{0.996}    \\ 
\hline
EPC++~\cite{luo2019every}                   & $640 \times 192$            & MS                     & 0.123          & 0.754          & 4.453          & 0.172          & 0.863           & 0.964             & 0.989             \\
Monodepth2~\cite{monodepth2}              & $640 \times 192$            & MS                     & { 0.080}    & { 0.466}    & { 3.681}    & { 0.127}    & { 0.926}     & { 0.985}       & { 0.995}       \\
BRNet~\cite{brnet}            & $640 \times 192$            & MS                     & {\ul 0.078} & {\ul 0.393} & {\ul 3.400} & {\ul 0.120} & {\ul 0.928}  & {\ul 0.988}    & \textbf{0.997}    \\
RPrDepth                   & $640 \times 192$            & MS                     & \textbf{0.068} & \textbf{0.341} & \textbf{3.212} & \textbf{0.105} & \textbf{0.946}  & \textbf{0.991}    & \textbf{0.997}    \\ \hline
\end{tabular}
}}
\caption{\textbf{Comparison on KITTI improved ground truth}. Comparison to other networks on 93\% KITTI 2015 Eigen split~\cite{eigen_split} and improve ground truth from~\cite{uhrig2017sparsity}.
}
\label{Table:improved}
\end{table*}

\begin{table*}[]
\centering
\small
\resizebox{1\textwidth}{!}{
\setlength{\tabcolsep}{1.5mm}{
\begin{tabular}{c|c|c|c|c|c|c|c|c|c|c}
\hline
\hline
\multirow{2}{*}{Method} & \multirow{2}{*}{Resolution} & \multirow{2}{*}{PostProcess} & \multirow{2}{*}{Train} & \multicolumn{4}{c}{lower is better}                                     & \multicolumn{3}{c}{higher is better}                       \\ \cline{5-11}
                        &                             &                              &                        & Abs Rel        & Sq Rel         & RMSE           & RMSE log             & $\delta < 1.25$ & $\delta^2 < 1.25$ & $\delta^3 < 1.25$    \\
\hline
Monodepth2~\cite{monodepth2}              & $640 \times 192$            &                              & M                      & 0.115          & 0.903          & 4.863          & 0.193                & 0.877           & 0.959             & 0.981                \\
Monodepth2~\cite{monodepth2}              & $640 \times 192$            &  \checkmark         & M                      & 0.112          & 0.851          & 4.754          & 0.190                & 0.881           & 0.960             & 0.981                \\
BRNet~\cite{brnet}                  & $640 \times 192$            &                              & M                      & { 0.105}    & { 0.698}    & { 4.462}    & { 0.179}          & { 0.890}     & {0.965}    & {0.984}       \\
BRNet~\cite{brnet}                  & $640 \times 192$            &  \checkmark         & M                      & {0.104} & {0.681} & {4.419} & {0.178}       & {0.891}  & {0.965}    & {0.984}       \\ 
RPrDepth                   & $640 \times 192$            &                              & M                      & {\ul 0.097}    & {\ul 0.658}    & {\ul 4.279}    & {\ul 0.169}          & \textbf{0.900}     & \textbf{0.967}    & \textbf{0.985}       \\
RPrDepth                   & $640 \times 192$            &  \checkmark         & M                      & \textbf{0.096} & \textbf{0.645} & \textbf{4.213} & \textbf{0.168}       & \textbf{0.900}  & \textbf{0.967}    & \textbf{0.985}       \\ 
\hline
Monodepth2~\cite{monodepth2}              & $640 \times 192$            &                              & S                      & 0.109          & 0.873          & 4.960          & 0.209                & 0.864           & 0.948             & 0.975                \\
Monodepth2~\cite{monodepth2}              & $640 \times 192$            &  \checkmark         & S                      & 0.108          & 0.842          & 4.891          & 0.207                & 0.866           & 0.949             & 0.976                \\
BRNet~\cite{brnet}                   & $640 \times 192$            &                              & S                      & { 0.103}    & { 0.792}    & { 4.716}    & { 0.197}          & { 0.876}     & { 0.954}       & {0.978}       \\
BRNet~\cite{brnet}                   & $640 \times 192$            &  \checkmark         & S                      & {0.102} & {0.774} & {4.679} & {0.196}       & {0.879}  & {0.955}    & {0.978}       \\ 
RPrDepth                   & $640 \times 192$            &                              & S                      & {\ul 0.098}    & {\ul 0.716}    & {\ul 4.538}    & {\ul 0.185}          & {\ul 0.885}     & {\ul 0.960}       & \textbf{0.980}       \\
RPrDepth                   & $640 \times 192$            &  \checkmark         & S                      & \textbf{0.097} & \textbf{0.709} & \textbf{4.498} & \textbf{0.184}       & \textbf{0.887}  & \textbf{0.961}    & \textbf{0.980}       \\ \hline
Monodepth2~\cite{monodepth2}              & $640 \times 192$            &                              & MS                     & 0.106          & 0.818          & 4.750          & 0.196                & 0.874           & 0.957             & 0.979                \\
Monodepth2~\cite{monodepth2}              & $640 \times 192$            &  \checkmark         & MS                     & 0.104          & 0.786          & 4.687          & 0.194                & 0.876           & 0.958             & 0.980                \\
BRNet~\cite{brnet}                  & $640 \times 192$            &                              & MS                     & { 0.099}    & { 0.685}    & { 4.453}    & { 0.183}          & { 0.885}     & { 0.962}       & {0.983}       \\
BRNet~\cite{brnet}              & $640 \times 192$            &  \checkmark         & MS                     & {0.098} & {0.671} & {4.418} & {0.178}       & {0.886}  & {0.963}    & {0.983}       \\ 
RPrDepth                   & $640 \times 192$            &                              & MS                     & {\ul 0.095}    & {\ul 0.638}    & {\ul 4.232}    & {\ul 0.169}          & {\ul 0.902}     & \textbf{0.970}       & \textbf{0.985}       \\
RPrDepth                   & $640 \times 192$            &  \checkmark         & MS                     & \textbf{0.094} & \textbf{0.615} & \textbf{4.183} & \textbf{0.167}       & \textbf{0.903}  & \textbf{0.970}    & \textbf{0.985}       \\ 
\hline
\hline
\end{tabular}
}}
\caption{\textbf{Results of RPrDepth on KITTI Eigen split with different supervision types and post process}. M means monocular videos only and S means stereo image pairs, and MS means both. The best two results are shown in bold and underlined, respectively.
}
\label{Table:results}
\vspace{-2mm}
\end{table*}

\section{Effective of Post-Processing}

The post-processing method in depth estimation, as introduced by \cite{monodepth}, enhances testing results. This technique processes each test image twice: first in its original form and then flipped. The results from the flipped image are then re-flipped and averaged with the original results to produce the final outcome. This approach has been proven to significantly improve accuracy, as noted in \cite{monodepth2}, \cite{poggi2018learning}, and \cite{monodepth}. Following the methodology of \cite{monodepth2}, we applied this post-process to our model in three different training settings and two resolutions.

As indicated in Table~\ref{Table:results}, applying post-processing results in noticeable gains for RPrDepth across all types of supervision and resolutions. Particularly, when RPrDepth is trained with Multi-Scale (MS) settings and used with a larger input resolution (640x192), it achieves impressive metrics of 0.094 in Absolute Relative (Abs Rel) and 4.183 in Root Mean Square Error (RMSE).

\begin{figure*}[t!]
  \centering
  \resizebox{1\textwidth}{!}{
  \includegraphics[width = 17cm]{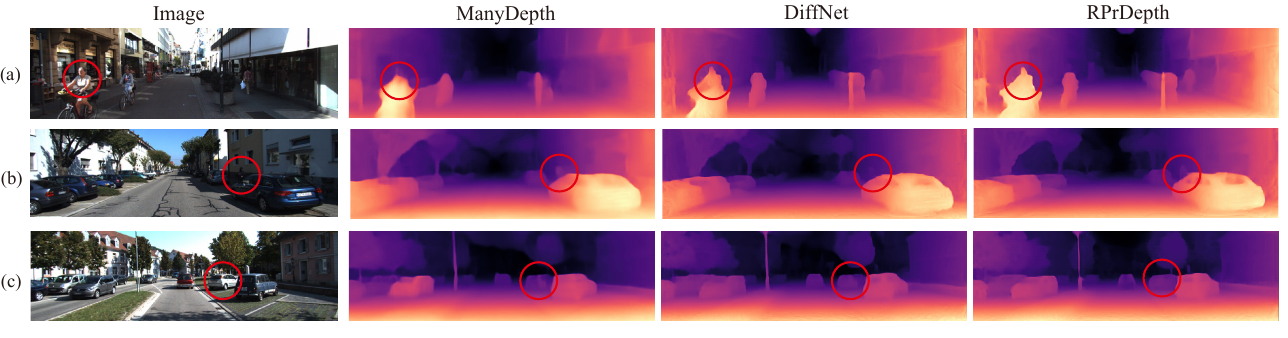}}
  \caption{\textbf{Additional qualitative results on the KITTI Eigen split test set.}}
  \vspace{-5mm}
  \label{Fig:qualitive}
\end{figure*} 

\section{Additional Qualitative Results}

For a clear comparison between RPrDepth and existing networks, additional qualitative results are showcased in Fig.~\ref{Fig:qualitive}. In this figure, we draw comparisons between RPrDepth, our baseline model DIFFNet \cite{zhou_diffnet}, and the guiding model ManyDepth \cite{manydepth}. The figure highlights that our method, RPrDepth, provides the most precise predictions when compared to the other methods. The most significant areas of difference are emphasized using red circles in the figure.
{\small
\bibliographystyle{ieee_fullname}
\bibliography{egbib}
}